\documentclass[lettersize,journal]{IEEEtran}
\usepackage{amsmath,amsfonts}
\usepackage{algorithmic}
\usepackage{algorithm}
\usepackage{xcolor}
\usepackage{array}
\usepackage[caption=false,font=normalsize,labelfont=sf,textfont=sf]{subfig}
\usepackage{textcomp}
\usepackage{stfloats}
\usepackage{url}
\usepackage{verbatim}
\usepackage{graphicx}
\usepackage{hyperref}
\usepackage{cite}
\usepackage{orcidlink}
\usepackage{multirow}
\begin{document}

\title{CoGReV: A Confidence-Gated Post-Hoc Non-Monotonic Belief Revision Framework for Phishing Website Classification} 
 
\author{Mainak Sen\orcidlink{0000-0001-5328-9248},
        Kumar Sankar Ray\orcidlink{0000-0002-3512-5810} and
        Amlan Chakrabarti\orcidlink{0000-0003-4380-3172}
        
\thanks{Mainak Sen(E-mail:mainak.s@technoindiaeducation.com, mainaksen.1988@gmail.com) is with Techno India University, West Bengal and is also a Ph.D scholar at the A. K. Choudhury School of Information Technology, University of Calcutta, Kolkata, India(msakc\_rs@caluniv.ac.in).\\
 Kumar Sankar Ray is with Techno India University, West Bengal, India, and is also with the Electronics and Communication Sciences Unit (ECSU), Indian Statistical Institute, Kolkata (E-mail: ksray@isical.ac.in). \\
Amlan Chakrabarti is with the A. K. Choudhury School of Information Technology, University of Calcutta, Kolkata, India 
(E-mail: acakcs@caluniv.ac.in) and is also a Visiting Professor with the Department of Artificial Intelligence, Indian Institute of Technology Kharagpur (E-Email: amlanc@ai.iitkgp.ac.in).\\
This work has been submitted to the IEEE for possible publication. Copyright may be transferred without notice, after which this version may no longer be accessible.
 }
 }  
\maketitle
\begin{abstract}
In phishing detection, machine learning classifiers act as a first line of defense, but the false positives they produce are triaged by human analysts. The excessive false alarms cause alert fatigue that erodes human oversight. We propose CoGReV, a hybrid framework that augments standard machine learning classifiers with a post-hoc non-monotonic reasoning layer implemented in Answer Set Programming. The layer applies a confidence-gated defeasible rule that revises a phishing prediction toward legitimate only when website metadata is present and the classifier's decision is low-confidence, deferring uncertain predictions to the reasoning layer while leaving out confident decisions to the classifiers. This gating acts as a function-allocation mechanism between the classifier and the reasoning layer. Unlike an ungated rule, which reduces false positives only by discarding genuine detections and degrades phishing recall by about nine percentage points, the proposed gated rule keeps recall within 0.7 percentage points of the no-revision baseline for all classifiers while revising only 0.27 percent of decisions. By lowering false alarms, without sacrificing detection, CoGReV reduces analysts alert fatigue and integrates new domain knowledge into the reasoning layer in $\mathcal{O}(n)$ time.

\end{abstract}

\section{Introduction}
\IEEEPARstart{P}{hishing}, is a prevalent form of cyber-attack that exploits social engineering techniques to deceive users into interacting with malicious web resources. These attacks involve deceptive techniques that trick users into revealing sensitive information like login credentials, sensitive data etc. 
The consequences of phishing attacks are severe, leading not only to financial losses 
but also to data breaches and reputational damage \cite{putra2024analysis}. 
To address the threat of phishing, researchers have created machine learning and deep-learning based URL classifiers
\cite{zara2024phishing}. These models are trained on static datasets to learn the inherent pattern in data and subsequently used as a frontline mechanism to combat phishing attacks. 
Once a classifier is trained and established, attackers try to fool the detection by using many obfuscation techniques\cite{marchal2014phishstorm} like swapping characters, typo-squatting etc. in different parts of URL. But, to identify and learn new attack pattern, the classifier needs to be trained again. So, dependency and trustworthiness on these classifiers is questionable as phishing URLs may pass learned patterns while still pointing to malicious destinations. 

Moreover, in operational security settings false positives are especially costly. Excessive false alarms cause alert fatigue \cite{ban2023breaking}, where analysts become desensitized and may overlook genuine threats. Phishing detection is inherently a human-machine task, in which the classifier makes automated decisions, but a human analyst remains responsible for reviewing flagged cases. Reducing false positives without sacrificing detection accuracy directly lowers the analyst's review burden which helps to sustain trust in the automated system. The objective is thus not only algorithmic accuracy, but a machine side   mechanism that supports the human operator by deferring only uncertain decisions to a reasoning layer while leaving confident decisions to the classifier.\\ 
 Meta tags in website provides valuable information such as description, author, keywords etc. which helps to identify legitimate details. Phishers try to hide these valuable details in case of malicious ones yet would try to make a URL look like a legitimate entry. Sometimes, attackers deliberately craft structural elements (meta data and content descriptors) to appear authentic while not reflecting true intent of the page\cite{ghalechyan2024phishing}. Thus, a deeper analysis of website's structural and contents based features provides additional contextual cues for reasoning. Real world environments like phishing detection are dynamic and often involve incomplete or evolving information. To address these limitations, human-like reasoning and belief revision are required for contextual decision support in phishing website classification. \\
 Non-monotonic reasoning (NMR) 
 \cite{brewka1997nonmonotonic}
 provides this capability by enabling a system to withdraw or revise conclusions when new or contradictory evidence is presented. Unlike conventional learning systems, which rely only on statistical correlation, NMR incorporates domain knowledge and common sense rules to logically reassess classifier decision. This is particularly useful in phishing detection, where subtle contextual clues may contradict initial patterns.\\
 In this work, we integrate machine learning based classifiers with NMR-based reasoning layer to form a dual framework phishing detection framework. The proposed approach is a belief revision based framework in which the first part performs initial classification of URL and the second part, NMR module evaluates these predictions against contextual knowledge to either confirm or alter them. 
 The proposed method employs Answer Set Programming (ASP) to perform non-monotonic reasoning over symbolic representations. Correcting the errors of a deployed classifier as attackers adapt conventionally requires retraining the model, which is costly and time consuming. The primary objective of PHISHREV is therefore not to maximize classification accuracy through alternate approaches, but rather to demonstrate how new contextual evidence and corrective knowledge can be incorporated into an existing classification system \emph{post-hoc} in $\mathcal{O}(n)$ time via post-hoc non-monotonic reasoning. The main contributions of this work are summarized as follows:
 \begin{enumerate}
 \item  A novel contextual binary feature \textit{meta\_info\_present} was constructed by parsing each URL and setting a flag for meta\_tag availability. 
 \item A confidence-gated post-hoc non-monotonic framework is proposed that formally revises initial classifier beliefs using Answer Set Programming. By deferring only low-confidence predictions to the reasoning layer, the rule reduces false positives while preserving phishing recall and integrates new domain knowledge in $O(n)$ time without retraining.  
 \end{enumerate}
\section{Related Work}
Security protocols can be categorized into software based approaches namely blacklist and whitelist \cite{das2019sok} based. Blacklist contains those which user should not visit while white-list contains all legitimate entries. 
Heuristic based detection mechanism depends on different features extracted from different parts of URL like lexical, content etc. 
Machine learning based classifiers uses extracted features from URL or website content and used for classification task. Support vector machine(SVM), random forest(RF), Naive Bayes(NB), J48, C4.5, k-Nearest Neighbor (kNN) are frequently used in this field\cite{odeh2021machine}. Various deep learning (DL)based architectures overcome the need of manual feature engineering as they can extract the features automatically from raw data. The convolutional neural network (CNN), deep neural network (DNN), recurrent neural network(RNN), Long Short-Term Memory (LSTM), Gated Recurrent Unit (GRU) and deep belief network (DBN) are frequently used architectures\cite{do2022deep} by researchers. 
Despite these advancements, existing approaches largely rely on statistical models and lack the ability to incorporate contextual evidence for adaptive decision refinement. These works address false positives typically by retraining the classifier (cost-sensitive learning) or on a global threshold adjustment, which act on model itself rather than correcting its outputs post-training. This motivated us to create a hybrid framework capable of post-hoc belief revision.
\section{Proposed Method}
\begin{figure}[!t]
\centering
\includegraphics[height=5cm, width=8cm]{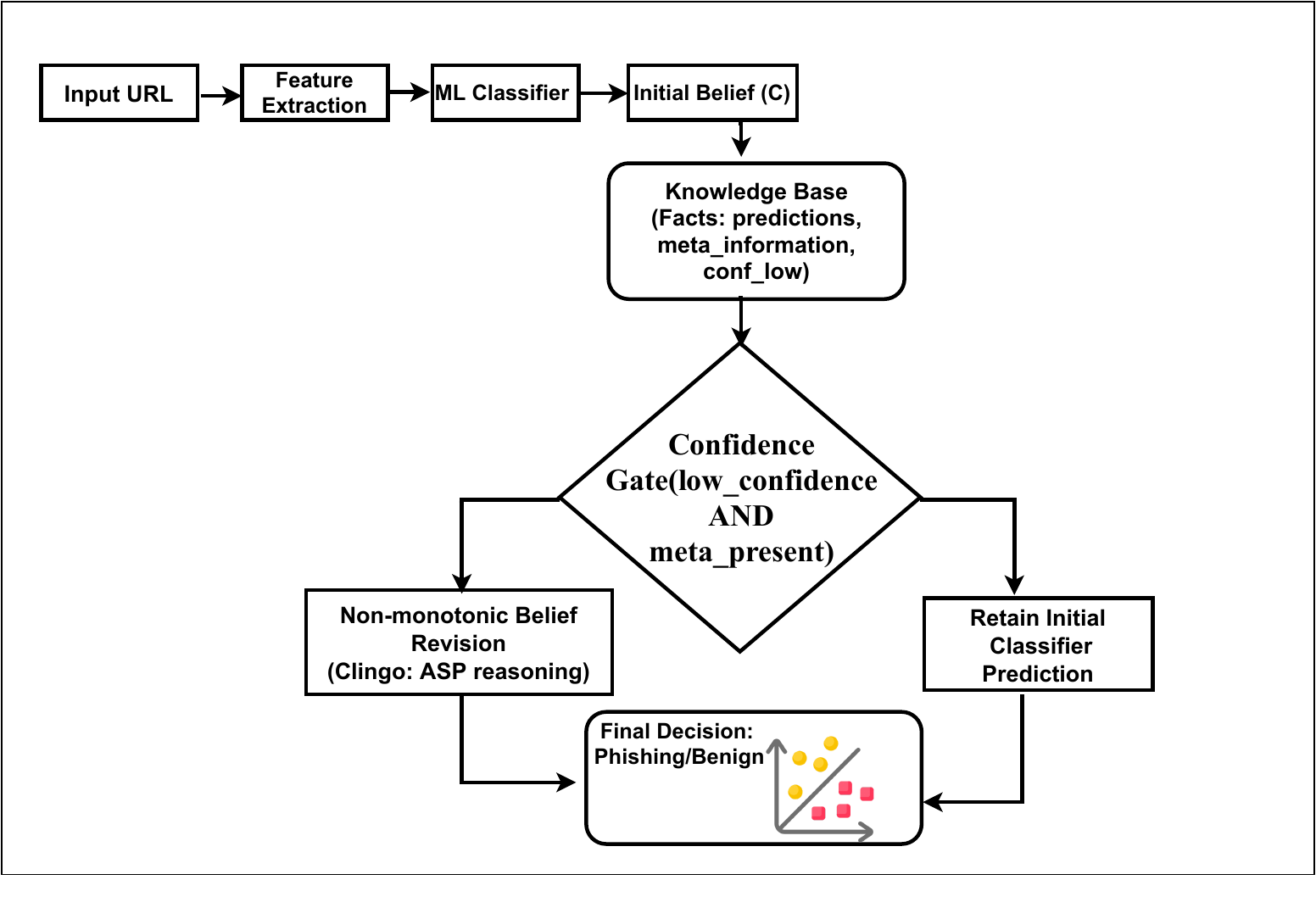}
\caption{Proposed hybrid phishing detection framework integrating machine learning and non-monotonic reasoning(NMR).}
\label{versionaugimage.png}
\end{figure}

The proposed dual framework works in two sequential phases, namely statistical prediction and post hoc symbolic reasoning. The whole procedure is depicted in figure \ref{versionaugimage.png}. 

\subsection{First Phase: Statistical Learning and Initial Belief Generation}
In the first phase, all 87 features present in the dataset are normalized using standard scaler technique to ensure consistent feature scaling. Each classifier produces an initial prediction  $C \in \{0,1\}$ where 0 represents legitimate and 1 means phishing URLs, together with a phishing class probability $P(CL, u_i)$. These predictions create the \textit{initial beliefs} about each instance. Formally, for each URL instance $u_i$ and classifier $CL \in \{SVM, KNN, DT, RF\}$, the initial belief $B_0(CL, u_i)$ is defined as the class label predicted by $CL$ from the lexical feature vector of $u_i$, the associated probability $P(CL, u_i)$ is retained for the confidence gate. 
\subsection{Second Phase: Post-hoc Non-Monotonic Reasoning Layer}
In this second stage, the initial beliefs generated by the classifiers are transformed into symbolic representations and incorporated into a knowledge base. Knowledge base is constructed at the instance level rather than aggregated statistics enabling fine grained reasoning over individual data points. The knowledge base consists of classifiers initial predictions and meta field availability or not for each URL. For each instance URL \textit{ID}, classifiers predictions are encoded as facts of the form \texttt{pred(CL, ID, C)}, while the availability metadata  is presented as \texttt{meta(ID, M)}, where $M \in \{\text{yes}, \text{no}\}$  where M=yes means meta\_available and M=no means meta\_not\_available. The non-monotonic reasoning component is implemented using Answer Set Programming(ASP) with Clingo\cite{GEBSER_KAMINSKI_KAUFMANN_SCHAUB_2019}. It enables post hoc symbolic reasoning over a knowledge base which consist of facts and rules. For each instance classifier also emits a phishing-class probability $P(\mathrm{CL},\mathrm{ID})$, and a prediction is labelled \emph{low-confidence} when it lies within a margin $\tau$ of the decision boundary , i.e.\
$0.5 < P(\mathrm{CL},\mathrm{ID}) \le 0.5 + \tau$. This condition is evaluated during fact generation and encoded as $\mathit{conf\_low}(\mathrm{CL},\mathrm{ID})$, alongside the $\mathit{pred}$ and $\mathit{meta}$ facts. Equation \ref{eq:guarded} is the confidence-gated defeasible rule used to revise the classifier belief. The belief is reversed to \emph{benign} only when the prediction is phishing, metadata is present, \emph{and} the prediction is low confidence, otherwise the classifier prediction is retained.

 \begin{equation}
\label{eq:guarded}
\text{Final Belief} =
\begin{cases}
\text{benign}, & \text{if } C{=}\text{phishing} \wedge M{=}\text{yes} \\
              & \quad \wedge\ \mathit{conf\_low}(\mathrm{CL},\mathrm{ID}) \\
C, & \text{otherwise}
\end{cases}
\end{equation}


Unlike conventional post-processing method that rely on statistical thresholds or probability calibration, equation \ref{eq:guarded} encodes a defeasible, confidence-conditioned revision rule within the ASP framework. The revision is non-monotonic in nature that is, a conclusion previously derived by the classifier is formally withdrawn upon the arrival of new contextual evidence, namely the presence of meta information. 
\subsubsection{Confidence Gate and Selection of the Margin $\tau$}\label{sec:tau}
Since metadata is a weak and spoofable feature, allowing it to overturn every phishing prediction is unsafe. We therefore, gate the revision by classifier confidence: metadata is permitted to revise a prediction only when the classifier was already uncertain about it (a low-confidence prediction), while confident predictions are left unchanged. A standard classifier collapses its output probability into a hard label via argmax, discarding the distinction between a confident prediction and a near-boundary one. Our confidence gate preserves this distinction: predictions close to the decision boundary $(low P(phishing))$ are treated as uncertain and deferred to the reasoning layer, rather than trusting firmly as high-confidence predictions. This acts as a function-allocation mechanism between the autonomous classifier and reasoning layer, so that only low confidence decisions are deferred for revision before reaching the human analyst.
Our revision layer is related to \emph{selective prediction}, or classification with a reject option. Selective prediction is a method \cite{el2010foundations, geifman2017selective} where a model abstains on input it is not confident about rather than committing to a decision. Selective classifiers trade coverage for reliability by acting only on high-confidence predictions and deferring the rest. Here, instead of abstaining, our framework routes low-confidence predictions to a symbolic non-monotonic reasoning layer, keeping high-confidence predictions are retained by the classifier. The margin $\tau$ is a hyperparameter selected on the training set only, using $5$-fold out-of-fold cross-validated predictions. For a grid of candidate margins we measure, on the training folds, the false positives removed, the false negatives introduced, and the resulting phishing-recall degradation. We impose a recall-preservation constraint and choose $\tau$ as in equation \ref{eq:tau}, the largest margin that keeps the worst-classifier training recall drop below half a percentage point. This yields $\tau= 0.12$, (training recall drop 0.44 pp; the next grid point $tau= 0.14$ exceeds $\epsilon$ at 0.61 pp), which is frozen and applied to test set.
\begin{equation}
\label{eq:tau}
\tau^{\star} = \max\{\, \tau : \Delta\mathrm{Recall}_{\mathrm{train}}(\tau) \le \epsilon \,\},
\qquad \epsilon = 0.5~\text{pp},
\end{equation}

\section{Dataset information}
We experimented on the dataset available at \cite{hannousse2021web} which contains total 11430 URLs. 
We independently parsed the respective web-pages and recorded the presence or absence of meta fields such as description, keyword and author.
\section{Experimental Results}
This section presents, the evaluation of both phases of the proposed framework- statistical learning and post-hoc belief revision.
\subsection{First Phase: Classifiers predictions- Initial Belief Generation}
\subsubsection{Hyper-parameters used for classification task}
We used four machine learning classifier with different hyper-parameters. For support vector machine, we explored different regularization strengths and kernel functions by using varying penalty parameter as $C \in \{0.1, 1, 10\}$, kernel types \{linear, rbf\} and kernel coefficient \{scale, auto\}. The k-NN classifier was tuned by number of neighbors $N \in \{3,5,7,9\}$, weighting strategies \{uniform, distance\} and distance function as \{euclidean, manhattan\}. For the decision tree classifier, we varied the splitting criterion \{gini, entropy\}, maximum tree depth \{3, 5, 10, None\} and minimum samples required to split an internal node \{2, 5, 10\}. Finally, random forest classifier, was configured with number of estimators \{100,  200\}, maximum depth \{5, 10, None\} and splitting criterion \{gini, entropy\}. 
The dataset was partitioned using 80:20 train-test split and model selection was performed by 5 fold cross validation on the training set. The results received by using different hyper-parameters for the experiment are placed in table \ref{tab:split_table_datasetB}.



\begin{table}[!t]
\centering
\caption{Performance Comparison of Classification Algorithms  (Legitimate(0), Phishing(1)}
\begin{tabular}{|p{1.38cm}|p{1.cm}|p{1cm}|p{.78cm}|p{1.07cm}|}
\hline
\textbf{Algorithm} & \textbf{Accuracy} & \textbf{Precision} & \textbf{Recall} & \textbf{F1-Score} \\
& (\%) & & & \\
\hline
SVM    & 96.32  & 0: 0.96 1: 0.96  & 0: 0.96  1: 0.96   & 0: 0.96   1: 0.96  \\
\hline
K-NN   & 95.18  & 0:  0.95 1: 0.96 & 0: 0.96 1: 0.94  & 0: 0.95 1: 0.95  \\
\hline
Decision Tree (DT) & 93.83 &  0: 0.94 1: 0.94 & 0: 0.94 1: 0.94  & 0: 0.94  1: 0.94   \\
\hline
Random Forest (RF)  & 95.89  & 0: 0.96 1: 0.96 & 0: 0.96 1: 0.96  & 0: 0.96  1: 0.96  \\
\hline
\end{tabular}
\label{tab:split_table_datasetB}
\end{table}

\subsubsection{Initial Belief Generation}
The test set had 2286 instances (20\%) whose predicted class labels are treated as initial beliefs for the second phase non-monotonic reasoning layer. For each test instance, in addition to the predicted label, we record each classifier's phishing-class probability $P(\mathrm{CL},\mathrm{ID})$, obtained from \texttt{predict\_proba} for KNN, DT, RF and from Platt-scaled  scores for SVM. These probabilities drive the confidence gate of Section~\ref{sec:tau}.

\subsubsection{Belief Revision based on Meta Tag}
Phishing websites often omit meta information like author, description, keywords etc. which is consistent with legitimate web development practice. Good or benign websites use meta information to support search engine ranking and enable semantic indexing. Motivated by this distinction, we incorporate the availability of meta information as an auxiliary feature for post-hoc belief revision. Metadata is treated as a supporting evidence in non-monotonic reasoning layer, where its presence can trigger the revision of phishing predictions generated by classifiers.  
In our dataset, from 5715 legitimate URLs, we found 2887 URLs which is approximately 50.52\% were having meta tags while in phishing class, 569 URLs out of 5715 that is 9.95\% of phishing URLs were having meta tags. Table \ref{tab:meta_distribution} presents the detail distribution of meta tags.

\begin{table}[h]
\centering
\caption{Meta Tag Distribution Across Train and Test Sets}
\label{tab:meta_distribution}
\begin{tabular}{|p{1.3cm}|p{.6cm}|p{1.7cm}|p{1.6cm}|p{1.0cm}|}
\hline
\multicolumn{5}{|c|}{\textbf{Training Set (80\% = 9,144 instances)}} \\
\hline
\textbf{URL Type} & \textbf{Total} & \textbf{Meta Present} & \textbf{Meta Absent} & \textbf{Meta(\%)} \\
\hline
Legitimate  & 4,572 & 2,319 & 2,253 & 50.72 \\
Phishing   & 4,570 &   457 & 4,113 & 10.00 \\
Total  & 9,143 & 2,776 & 6,367 & 30.37 \\
\hline
\multicolumn{5}{|c|}{\textbf{Test Set (20\% = 2,286 instances)}} \\
\hline
\textbf{URL Type} & \textbf{Total} & \textbf{Meta Present} & \textbf{Meta Absent} & \textbf{Meta(\%)} \\
\hline
Legitimate  & 1,143 & 568 &   575 & 49.69\\
Phishing & 1,143 & 112 & 1,031 &  9.80 \\
Total          & 2,286 & 680 & 1,606 & 29.75 \\
\hline
\end{tabular}
\end{table}

\subsection{Second Phase: Post-hoc Non-Monotonic Reasoning Layer}
We used non-monotonic reasoning  to revise the initial classifier prediction. Based on the domain specific rules, the reasoning layer selectively revises certain predictions, enabling belief updates. 
Table \ref{tab:revisions} states the count of samples whose belief was altered by NMR module. With the confidence gate active at $\tau=0.12$, the reasoning layer revises $25$ predictions in total across all the four classifiers. The revisions are concentrated on the classifiers whose predictions lie close to the decision boundary, with Random Forest accounting for $18$ of the $25$. For the well calibrated SVM, the gate revises only a single prediction, correctly declining to overturn its confident decisions. Overall, out of 9,144(20\% test set) classifier level  predictions, $25$ were revised by the confidence-gated non-monotonic reasoning layer which is approximately $0.27\%$ of all decisions. This demonstrates that the proposed post-hoc non-monotonic reasoning layer performs selective, evidence driven belief revision. \\
Table~\ref{tab:main_results} presents the comparison of false positives (FP) across all four classifiers before and after the application of the NMR-based belief revision. Aggregated over the four classifiers, the gated rule lowers the total false positives from $206$ to $191$, that is, it reduces $15$ false alarms and directly addresses alert fatigue\cite{ban2023breaking}. Over the same predictions, the total false negatives rise only from  $223$ to $233$, i.e just $10$ phishing sites are missed. The phishing catch rate is essentially preserved that is recall stays within $0.7$\ percentage points of the baseline for every classifier. Precision improves for the classifiers whose false positives fall(e.g.\ Random Forest, $0.9573 \rightarrow 0.9655$), reflecting cleaner phishing alerts. From a human-machine standpoint, each false positive reduction is one fewer benign alert an analyst must review. 
\begin{table}[!t]
\centering
\caption{Predictions Revised by the Confidence-Gated Non-Monotonic Reasoning Layer ($\tau=0.12$).}
\label{tab:revisions}
\begin{tabular}{|l|c|}
\hline
\textbf{Classifier} & \textbf{Revised Predictions}\\
\hline
SVM            & 1\\
KNN            & 5\\
Decision Tree  & 1\\
Random Forest  & 18\\
\hline
\textbf{Total} & \textbf{25}\\
\hline
\end{tabular}
\end{table}

\begin{table}[!t]
\centering
\caption{Effect of Confidence-Gated Non-Monotonic Revision ($\tau=0.12$) on the Test Set. Values are shown as \textit{before}$\rightarrow$\textit{after} revision.}
\label{tab:main_results}
\resizebox{\columnwidth}{!}{%
\begin{tabular}{|l|c|c|c|c|c|}
\hline
\textbf{Classifier} & \textbf{Rev.} & \textbf{FP} & \textbf{FN} & \textbf{Recall} & \textbf{Precision}\\
\hline
SVM           & 1  & $41\!\rightarrow\!41$ & $43\!\rightarrow\!44$ & $0.9624\!\rightarrow\!0.9615$ & $0.9641\!\rightarrow\!0.9640$\\
KNN           & 5  & $47\!\rightarrow\!43$ & $63\!\rightarrow\!64$ & $0.9449\!\rightarrow\!0.9440$ & $0.9583\!\rightarrow\!0.9617$\\
Decision Tree & 1  & $69\!\rightarrow\!68$ & $72\!\rightarrow\!72$ & $0.9370\!\rightarrow\!0.9370$ & $0.9395\!\rightarrow\!0.9403$\\
Random Forest & 18 & $49\!\rightarrow\!39$ & $45\!\rightarrow\!53$ & $0.9606\!\rightarrow\!0.9536$ & $0.9573\!\rightarrow\!0.9655$\\
\hline
\textbf{Total} & \textbf{25} & $\mathbf{206\!\rightarrow\!191}$ & $\mathbf{223\!\rightarrow\!233}$ & -- & --\\
\hline
\end{tabular}}
\end{table}

\subsection{Ablation Analysis}
\begin{figure*}[!t]
\centering
\includegraphics[height=5.1cm, width=17cm]{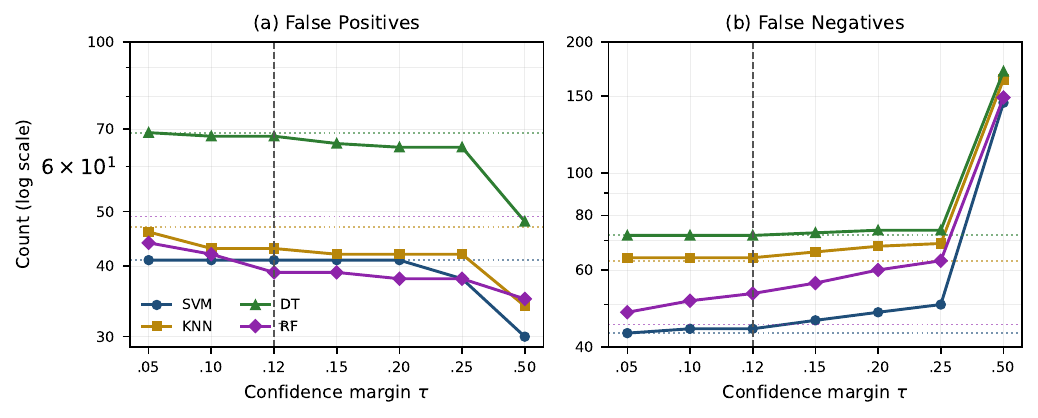}
\caption{Per-classifier false positives (a) and false negatives (b) versus the confidence margin $\tau$ (log scale). Solid curves show the gated rule; dotted lines of the same colour mark each classifier's no-revision baseline. The dashed vertical line is the selected operating point $\tau^\star=0.12$. As $\tau$ increases, false positives fall at or below baseline while false negatives stay near baseline until the ungated case ($\tau=0.5$), where they rise sharply for every classifier.}
\label{result}
\end{figure*}

The value of the confidence gate is made explicit in
Fig.~\ref{result}. It shows how false positives and false negatives evolve as the confidence margin  $\tau$ increases. At the ungated extreme ($\tau=0.5$), the rule revises $465$ predictions and lowers the total false positives from $206$ to $147$, but it inflates the total false nagatives from $223$ to $629$. In contrast, the selecting operating point $\tau^\star=0.12$ limits the false negative increse to only $10$ (from $223$ to $233$), removing approximately $98\%$ of this false negative cost while retaining a meaningful share of the false-positive reduction. Here, baseline is the classifier output with no revision applied. Fig.~\ref{result} shows, each classifier stays close to this baseline across the operating range. The false negatives which rises sharply at $\tau=0.5$ barely move. The gate therefore keeps recall intact while still cutting false positives.


\section{Comparison with state of art}
Table \ref{tab:sota} compares PHISHREV against existing phishing website detection studies that used the same benchmark. While prior works, focus exclusively on maximizing classification accuracy through monotonic statistical learning, our proposed PHISHREV uniquely augments machine learning predictions with a post-hoc ASP-based non-monotonic reasoning layer to reduce false positives.
\begin{table}[h]
\centering
\caption{Comparison with State-of-the-Art Methods}
\label{tab:sota}
\resizebox{\columnwidth}{!}{%
\begin{tabular}{|p{1.8cm}|p{1.8cm}|p{1.8cm}|p{1.8cm}|p{1.8cm}|}
\hline
\textbf{Metric} & 
Hannousse et al.~\cite{hannousse2021towards} & 
Adane et al.~\cite{adane2023single} & 
Shafin et al.~\cite{shafin2025explainable} & 
\textbf{PHISHREV} \\
\hline
RF Accuracy  & 
96.61 \% & 
96.50 \%& 
96.76 \% & 
95.89\% \\
\hline
Nature & 
Monotonic & 
Monotonic & 
Monotonic & 
Non-monotonic \\
\hline
Post-hoc ASP Reasoning & 
No & 
No & 
No & 
Yes \\
\hline
Training Complexity & 
$\mathcal{O}(T \!\cdot\! n \!\cdot\! f \!\cdot\! \log n)$ & 
$\mathcal{O}(T \!\cdot\! n \!\cdot\! f \!\cdot\! \log n)$ & 
$\mathcal{O}(T \!\cdot\! n \!\cdot\! f \!\cdot\! \log n)$ & 
$\mathcal{O}(T \!\cdot\! n \!\cdot\! f \!\cdot\! \log n)$ \\
\hline
NMR Overhead & 
Not Applicable & 
Not Applicable & 
Not Applicable & 
$\mathcal{O}(n)$ \\
\hline
Retraining Complexity & 
$\mathcal{O}(T \!\cdot\! n \!\cdot\! f \!\cdot\! \log n)$ & 
$\mathcal{O}(T \!\cdot\! n \!\cdot\! f \!\cdot\! \log n)$ & 
$\mathcal{O}(T \!\cdot\! n \!\cdot\! f \!\cdot\! \log n)$ & 
\textbf{Not Required} \\
\hline
\end{tabular}}
\end{table}

\section{Computational Complexity}
In the first phase, the ML classifiers are trained on the feature set(f) over n training samples. The training complexity of random forest (the best classifier) with T trees requires $\mathcal{O}(T \cdot n \cdot f 
\cdot \log n)$. In the second phase, the reasoning layer computes the confidence gate and generates the ASP facts from the classifier predictions in $O(n)$ time  and Clingo evaluates the revision rules over the fact base also in $\mathcal{O}(n)$ time. The advantage of the proposed method is that new domain knowledge such as emerging phishing evasion strategies or any legitimacy up-dation signal can be incorporated directly into the ASP knowledge base, without any need to retrain the underlying ML classifiers. 
\section{Discussion}
URL based detection system remains vulnerable to adversarial manipulation, as attackers can  modify URL structures, hosting strategies and meta information to evade detection. This highlights the need of additional information beyond purely lexical features, which motivates our use of metadata as an auxiliary contextual signal. A practical limitation is that many URLs in public phishing datasets become inactive once reported, which restricts the extraction of meta tags.\\ 
Metadata is a weak and spoofable cue. Phishers mimic the 
behaviour of trusted websites by injecting descriptive meta fields to deceive classifiers and in our dataset  569 phishing URLs (9.95\%) indeed carry meta tags. A revision rule that treated metadata presence as evidence of legitimacy would therefore overturn genuine detections and inflate false negatives. By using confidence gate, we address this risk directly. Metadata is admitted as evidence only for low-confidence phishing predictions, so it can overturn a borderline decision but never a confident detection. This is why the gated rule limits the false-negative increase to $10$, whereas the ungated rule adds more than $400$ (figure \ref{result}). 
\section{Conclusion and Future Scope}
This work presents a dual framework that integrates machine learning classifier with ASP-based non-monotonic reasoning for post-hoc belief revision in phishing detection system. Rather than revising predictions unconditionally, the framework applies a confidence-gated rule that defers only low-confidence predictions to the reasoning layer. This reduces false positives while keeping phishing recall within $0.7$ percentage points of the baseline and allows new domain knowledge to be incorporated in $O(n)$ time without retraining. 
\\The current framework relies on a single metadata-based rule, which may still yield incorrect revisions when an adversary supplies convincing metadata to a low-confidence phishing page. Incorporating additional contextual evidence through a multi-rule reasoning strategy is the direction for future work.

\section{Data Availability}
The notebook file, .lp files, altered instances files will be made available upon acceptance at GitHub repository. 

\bibliographystyle{IEEEtran}
\bibliography{Letter.bib}

\end{document}